\documentclass[runningheads]{llncs}
\usepackage{amsmath}
\usepackage{amsfonts}
\usepackage[T1]{fontenc}
\usepackage{graphicx,verbatim}
\usepackage{booktabs}
\usepackage{multirow}
\usepackage{makecell} % For multi-line column headers
\usepackage{tabularx, booktabs}
\usepackage[hidelinks,colorlinks=true,linkcolor=blue,citecolor=blue]{hyperref}
\usepackage{adjustbox}

\usepackage{units}

\usepackage{booktabs}
\usepackage{tabularx}
\usepackage{pdflscape}
\usepackage{array}
\begin{document}
\title{Federated Fine-tuning of SAM-Med3D for MRI-based Dementia Classification}

\titlerunning{Federated Fine-Tuning of SAM-Med3D for Dementia Classification} % Short version of the title 

\author{Kaouther Mouheb\thanks{Corresponding author: k.mouheb@erasmusmc.nl}\inst{1} \and
Marawan Elbatel \inst{2} \and
Janne Papma \inst{3} \and 
Geert Jan Biessels \inst{4} \and 
Jurgen Claassen \inst{5} \and 
Huub Middelkoop \inst{6} \and 
Barbara van Munster \inst{7} \and
Wiesje van der Flier \inst{8} \and 
Inez Ramakers \inst{9} \and 
Stefan Klein \inst{1} \and 
Esther E. Bron \inst{1}}
\authorrunning{K. Mouheb et al.} 
\institute{
Dept. of Radiology \& Nuclear Medicine, Erasmus MC, Rotterdam, the Netherlands \and
The Hong Kong University of Science and Technology, Hong Kong SAR \and
Dept. of Neurology, Erasmus MC, Rotterdam, the Netherlands \and
Dept. of Neurology, UMC Utrecht, Utrecht, the Netherlands \and
Dept. of Geriatrics, Radboud UMC, Nijmegen, the Netherlands \and
Dept. of Neurology, Leiden UMC, Leiden, the Netherlands \and
Dept. of Internal Medicine, UMC Groningen, Groningen, the Netherlands \and
Dept. of Neurology, Amsterdam UMC location VUmc, Amsterdam, the Netherlands \and
Dept. of Psychiatry \& Psychology, Maastricht UMC, Maastricht, the Netherlands
} 
\maketitle

\begin{abstract}
While foundation models (FMs) offer strong potential for AI-based dementia diagnosis, their integration into federated learning (FL) systems remains underexplored. In this benchmarking study, we systematically evaluate the impact of key design choices: classification head architecture, fine-tuning strategy, and aggregation method, on the performance and efficiency of federated FM tuning using brain MRI data. Using a large multi-cohort dataset, we find that the architecture of the classification head substantially influences performance, freezing the FM encoder achieves comparable results to full fine-tuning, and advanced aggregation methods outperform standard federated averaging. Our results offer practical insights for deploying FMs in decentralized clinical settings and highlight trade-offs that should guide future method development.
% These insights provide practical guidance for the deployment of FMs in real-world FL scenarios.

% Authors must provide keywords and are not allowed to remove this Keyword section.
\end{abstract}

\section{Introduction}

The accurate and early diagnosis of dementia is crucial for effective intervention and care \cite{ad_why_get_checked}. Training AI models for this task requires diverse, multi-center datasets to capture patient variability. However, centralizing such data raises significant privacy concerns \cite{murdoch2021privacy}. Federated learning (FL) addresses this challenge by allowing collaborative training while preserving data privacy \cite{Lei2024HybridFL,Huang2021FederatedLV}. However, FL faces challenges such as inter-client heterogeneity, which can hinder model convergence and performance \cite{Babar2024InvestigatingTI}. 
Foundation models (FMs) are large-scale models pre-trained on extensive datasets, capable of generalizing across diverse tasks. Integrating FMs into FL offers a promising approach to improve performance, as they act as powerful feature extractors, allowing efficient transfer learning for down-stream tasks \cite{bommasani2021opportunities,baharoon2023general}. Fine-tuning FMs in federated settings involves critical design decisions, including the selection of the classification head, fine-tuning strategy, and aggregation method. While prior research has explored some of these aspects in medical image segmentation \cite{liu2024fedfms} and 2D classification \cite{alkhunaizi2024probing}, their impact on 3D medical image classification remains unexplored.
This gap underscores the need for systematic investigation to optimize federated FM fine-tuning in this domain.

Transfer learning from FMs have shown strong promise in medical imaging. Baharoon et al. found that DINOv2, a general-purpose 2D FM, shows high performance in various medical tasks, including MRI-based classification \cite{baharoon2023general,oquab2023dinov2}. Wang et al. built SAM-Med3D, an FM trained fully on 3D medical images using a multimodal dataset of 140,000 scans \cite{wang2023sammed3d}.  In dementia research, Xue et al. built a multimodal FM for dementia diagnosis using public datasets \cite{xue2024ai}, building on the SwinUNETR model \cite{tang2022self}. 
The intersection of FL and FMs in medical tasks is gaining attention \cite{Jiang2024PrivacyPreservingFF}. For instance, Rate-My-LoRA was proposed to fine-tune FMs for cardiac MRI segmentation \cite{he2025rate}.
demonstrating the potential of FL to enhance 3D medical FMs 

Comparing these studies highlights key design choices that can impact model performance and efficiency. For example, Xue et al. integrated a convolutional adapter on top of Swin-UNETR, while Baharoon et al. used a simple linear layer as a classification head \cite{xue2024ai,baharoon2023general}. Fine-tuning methods vary, with some freezing the FM backbone \cite{xue2024ai}, while others use parameter-efficient fine-tuning (PEFT) methods such as low-rank adaptation (LoRA) to reduce communication overhead \cite{hu2021lora,he2025rate}. Aggregation methods also differ, from traditional algorithms such as FedAvg \cite{mcmahan2017communication}, to more advanced methods such as Rate-My-LoRA \cite{he2025rate}.
Existing work on federated FMs mainly focuses on 2D modalities and segmentation tasks. Moreover, many use simulated federations due to limited multi-center datasets, raising concerns about their clinical relevance and viability in medical practice. Thus, the impact of the different design choices on the performance and efficiency of federated FM fine-tuning for dementia diagnosis is poorly understood, highlighting the need for a rigorous and systematic evaluations on real-world multi-center datasets. 

In this work, we present a comprehensive empirical study on federated fine-tuning of a 3D FM (SAM-Med3D) for MRI-based dementia diagnosis. 
Our key contributions are:
(i) We develop an open-source framework for evaluating federated fine-tuning of 3D FMs in medical image classification.
(ii) We conduct a systematic analysis of three key design factors: classification head architecture, fine-tuning strategy, and federated aggregation technique, demonstrating their impact on diagnostic performance and efficiency.
(iii) We benchmark the methods on a large dataset of 6076 samples from multiple cohorts of diverse sources, offering actionable insights into deploying federated FMs in clinical settings.

\section{Materials and Method}
\subsection{Problem Formulation}
We aim to fine-tune an FM for a classification task using FL. The training data is distributed across a number of clients $N$, with the $i$-th client having a local dataset $\mathcal{D}_i = \{(\mathbf{x}_{i,j}, y_{i,j})\}_{j=1}^{n_i}$ where $\mathbf{x}_{i,j}$ represents the input 3D scan, and $y_{i,j} \in \{1, 2, \ldots, C\}$ is the corresponding class label. The model $f(\mathbf{x}; \boldsymbol{\theta})$ comprises two main components: a pre-trained FM that serves as a feature extraction backbone $g(\mathbf{x}; \boldsymbol{\theta}_g)$ and a classification head $h(\mathbf{z}; \boldsymbol{\theta}_h)$, where $\mathbf{z} = g(\mathbf{x}; \boldsymbol{\theta}_g)$ is the feature representation. The overall model is expressed as $f(\mathbf{x}; \boldsymbol{\theta}) = h(g(\mathbf{x}; \boldsymbol{\theta}_g); \boldsymbol{\theta}_h)$, with $\boldsymbol{\theta} = \{\boldsymbol{\theta}_g, \boldsymbol{\theta}_h\}$ denoting the model parameters. In FL, the model is trained at each client. The resulting local models \( f_i(\mathbf{x}; \boldsymbol{\theta}_i) \) are aggregated on the server into a global model \( f(\mathbf{x}; \boldsymbol{\theta}_{\text{glob}}) \), computed as a weighted average of the local models:
$\theta_{\text{glob}} = \sum_{i=1}^{N} \omega_i \boldsymbol{\theta}_i$
where \( \omega_i \) denotes the aggregation weight for client \(i\), determined by the aggregation method. The global model is sent back to the clients for the next round. This process is repeated for a number of rounds $R$. 

\begin{figure}[t]
\includegraphics[width=\textwidth]{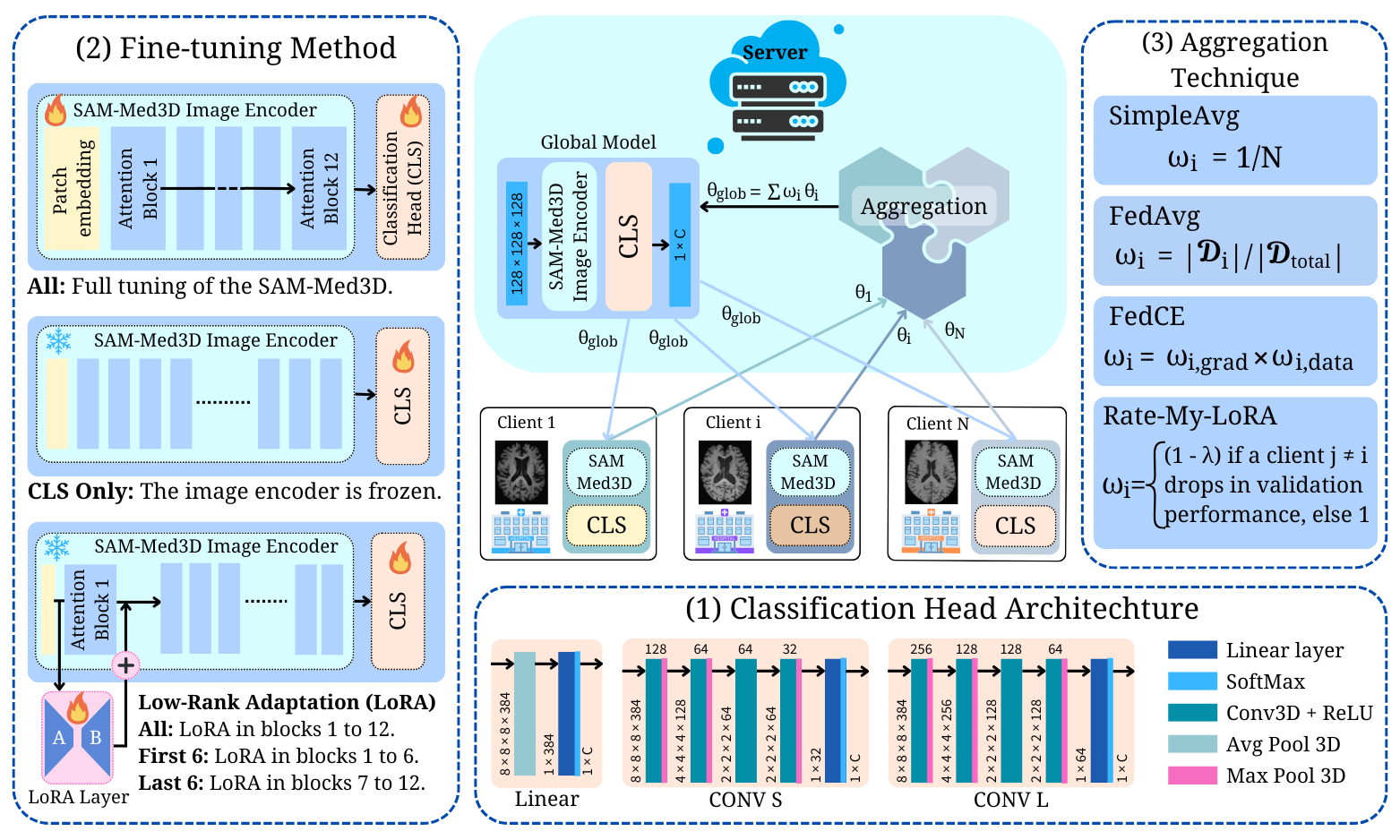}
\caption{The framework explores three design choices: (1) Classification Head Architecture: linear, small CNN adapter (CONV S), and large CNN adapter (CONV L); (2) Fine-tuning Method: full model tuning, classifier-only (linear probing), and LoRA with selective attention block adaptation; (3) Aggregation Strategy: including SimpleAvg, FedAvg, and the advanced methods FedCE and Rate-My-LoRA.} \label{fig:framework}
\end{figure}

\subsection{Evaluation Framework}
We identify three design choices that can impact performance and efficiency: the classification head architecture, the fine-tuning technique, and the aggregation method. To systematically assess their impact, each element is evaluated independently under consistent conditions. SAM-Med3D's image encoder \cite{wang2023sammed3d} is used as the backbone $g(\mathbf{x}; \boldsymbol{\theta}_g)$. This choice is motivated by the 3D nature of the model and its large medical training set. The framework is illustrated in Fig. \ref{fig:framework}.
% Below, we detail the experimental settings used to evaluate each element.
\textbf{Classification head architecture:}
The output of SAM-Med3D's encoder is 384 feature maps of shape ($8\times8\times8$). To classify samples based on this output, we evaluate three classification head architectures: (1) \textit{Linear:} an average pooling layer followed by a linear layer; (2) \textit{CONV S:} a lightweight 4-layer convolutional block with 128, 64, 64, and 32 kernels, followed by a linear layer; and (3) \textit{CONV L:} a 4-layer convolutional block with 256, 128, 128, and 64 kernels, followed by a linear layer. These architectures are selected based on prior work \cite{baharoon2023general,xue2024ai} to explore a trade-off between representational capacity and efficiency.\\
\textbf{Fine-tuning method:}
We compare 3 methods: (1) \textit{Full}: tuning all parameters in the model. (2) \textit{CLS Only} (linear probing): freezing the backbone $g(\mathbf{x}; \boldsymbol{\theta}_g)$ and training only the classifier $h(\mathbf{z}; \boldsymbol{\theta}_h)$.  (3) \textit{LoRA}: a technique that reduces the number of parameters to be trained. Let $\boldsymbol{\theta}_g$ denote the pre-trained parameters of the backbone $g(\mathbf{x}; \boldsymbol{\theta}_g)$. Suppose a linear layer in the encoder has a weight matrix $\mathbf{W} \in \mathbb{R}^{d \times k}$. LoRA introduces a trainable low-rank update $\Delta \mathbf{W} \in \mathbb{R}^{d \times k}$ as:
\[
\mathbf{W}_{\text{LoRA}} = \mathbf{W} + \Delta \mathbf{W} = \mathbf{W} + \mathbf{A} \mathbf{B}, \quad \mathbf{A} \in \mathbb{R}^{d \times r}, \quad \mathbf{B} \in \mathbb{R}^{r \times k}, \quad r \ll \min(d, k)
\]
where only the weights $\mathbf{A}$ and $\mathbf{B}$ are trainable. The backbone function becomes: $g_{\text{LoRA}}(\mathbf{x}) = g(\mathbf{x}; \boldsymbol{\theta}_g, \Delta \boldsymbol{\theta}_g)$, where $\Delta \boldsymbol{\theta}_g$ consists of the low-rank parameters $\{\mathbf{A}^{(l)}, \mathbf{B}^{(l)}\}_{l \in \mathcal{L}}$ for a subset of layers $\mathcal{L}$ to which LoRA is applied. We test configurations where LoRA is applied to all, the first 6, or the last 6 attention blocks of the encoder. We focus on linear probing and LoRA as they are the most commonly used PEFT methods, known for preserving pre-trained representations while reducing computational cost, making them well-suited for FL \cite{dutt2023parameter}.\\
\textbf{Aggregation technique:}
We evaluate two traditional aggregation methods, \textit{simple averaging}, which assigns equal weights to all clients ($\omega_i = 1/N$), and \textit{FedAvg} which weights clients based on their dataset size ($\omega_i = \nicefrac{|\mathcal{D}_i|}{\sum_{j=1}^{N} |D_j|}$). Furthermore, we explore two advanced methods: (1) \textit{FedCE} \cite{jiang2023fair}: the aggregation weight is given as $\omega_i =  \omega_i^{\text{grad}} \times \omega_i^{\text{data}},$ where $\omega_i^{\text{grad}}$ measures the alignment of client $i$'s gradient with those of other clients, reflecting its contribution in gradient space, and $\omega_i^{\text{data}}$ is the validation error of the model obtained by aggregating all clients excluding client $i$. A higher error is assumed to reflect greater contribution in the data space. (2) \textit{Rate-My-LoRA} \cite{he2025rate}: validation performance is monitored and higher weights are assigned to clients whose performance declines from the previous round.
FedCE and Rate-My-LoRA were selected because they are designed to address client heterogeneity in 3D medical imaging, and Rate-My-LoRA specifically addresses federated fine-tuning of 3D FMs.

\subsection{Implementation Details}
We used Nvidia-Flare and MONAI \cite{roth2022nvidia,cardoso2022monai}, with training distributed over 4 H100 GPUs. MRI scans were registered to the MNI template and skull-stripped \cite{fischer2003flirt,smith2000bet} and resized to $128^3$ (SAM-Med3D's input size). Models were trained for $R=10$ rounds with a batch size of 8, learning rate of 0.001 and the AdamW optimizer. LoRA rank was set to $r=8$, which outperformed other tested values ($r=4,16$). The code is available at \url{gitlab.com/radiology/neuro/fedmedsam_ad}.

\subsection{Baselines}
We compare against three baselines: (1) a 3D CNN (ResNet18) trained from scratch using FedAvg, following current FL practices for dementia diagnosis \cite{guan2024federated}; (2) centralized fine-tuning with a frozen encoder and ‘CONV S’ classifier; and (3) a nearest centroid classifier (NCC) using frozen encoder features, where clients share class-wise feature sums and counts. The global centroid for class $c$ is $\mu_c = \frac{1}{n_c} \sum_{i=1}^{N} \sum_{j:y_{i,j} = c} g(\mathbf{x}_{i,j}; \boldsymbol{\theta}_g)$, with $n_c$ the total number of samples in class $c$.

\subsection{Datasets}
We compiled a large dataset of 6,076 brain MRI scans from multiple sources reflecting a realistic and heterogeneous setting. The dataset consists of the Alzheimer's Disease Neuroimaging Initiative (ADNI) \cite{mueller2005alzheimer}, a clinical cohort of the Open Access Series of Imaging Studies (OASIS-4) \cite{marcus2007open}, the Neuroimaging in Frontotemporal Dementia Study (NIFD) \cite{ftldni}, the National Alzheimer's Coordinating Center (NACC) cohort \cite{beekly2004nacc}, the Latin American Brain Health Institute dataset (BrainLAT) \cite{prado2023brainlat}, as well as the Health-RI Parelsnoer Neurodegenerative Diseases Biobank (PND) \cite{aalten2014dutch} which consists of data acquired from 8 medical centers in the Netherlands. These sources cover a wide range of dementia subtypes, geographical location and demographic variability, providing a robust benchmark for evaluating dementia diagnosis models in a federated setting. In our experiments, each cohort is treated as a client in the federation. The task is to classify dementia patients (DE) and cognitively normal (CN) individuals. Subjects without T1-weighted brain MRI scans were excluded from the analysis. Table~\ref{tab:datasets} shows the label distribution per client. To illustrate inter-client variability in image appearance, we provide intensity histograms (Fig. S1, Appendix).
% computed from all voxels across all scans in each client’s dataset.

% the Health-RI Parelsnoer Neurodegenerative Diseases Biobank (PND) \cite{aalten2014dutch}, 

\begin{table}[t]
\centering
\caption{Diagnosis distribution per client for the train, validation and test splits}
\begin{adjustbox}{max width=\textwidth}
\begin{tabularx}{\textwidth}{l!{\vrule width 0.5pt}>{\centering\arraybackslash}X>{\centering\arraybackslash}X!{\vrule width 0.5pt}>{\centering\arraybackslash}X>{\centering\arraybackslash}X!{\vrule width 0.5pt}>{\centering\arraybackslash}X>{\centering\arraybackslash}X!{\vrule width 0.5pt}>{\centering\arraybackslash}X>{\centering\arraybackslash}X!{\vrule width 0.5pt}>{\centering\arraybackslash}X>{\centering\arraybackslash}X!{\vrule width 0.5pt}>{\centering\arraybackslash}X>{\centering\arraybackslash}X!{\vrule width 0.5pt}>{\centering\arraybackslash}X}
% \hline
Client & \multicolumn{2}{c!{\vrule width 0.5pt}}{ADNI} & \multicolumn{2}{c!{\vrule width 0.5pt}}{NIFD} & \multicolumn{2}{c!{\vrule width 0.5pt}}{OASIS} & \multicolumn{2}{c!{\vrule width 0.5pt}}{NACC} & \multicolumn{2}{c!{\vrule width 0.5pt}}{BrainLAT} & \multicolumn{2}{c!{\vrule width 0.5pt}}{PND} & Total \\
Label  & DE & CN & DE & CN & DE & CN & DE & CN & DE & CN & DE & CN & \\
\hline
Training & 240 & 516 & 98 & 74 & 224 & 28 & 683 & 1262 & 210 & 106 & 119 & 82 & 3642 \\
Validation & 40 & 86 & 17 & 12 & 37 & 5 & 113 & 211 & 35 & 18 & 20 & 13 & 607 \\
Test & 121 & 258 & 49 & 38 & 113 & 14 & 342 & 632 & 105 & 54 & 60 & 41 & 1827 \\
% \hline
\end{tabularx}
\end{adjustbox}
\label{tab:datasets}
\end{table}

\section{Experimental Results}
We assess diagnostic performance using AUC, communication efficiency via average message size and latency, and computational efficiency via GPU memory, energy usage, and FLOPs per sample. AUC per client is reported in Table S.1. We compute 95\% confidence intervals (CI) from 10,000 test-set bootstraps; non-overlapping CIs indicate statistical significance.
\subsection{Classification Head Architecture}
\begin{figure}[t]
\includegraphics[width=\textwidth]{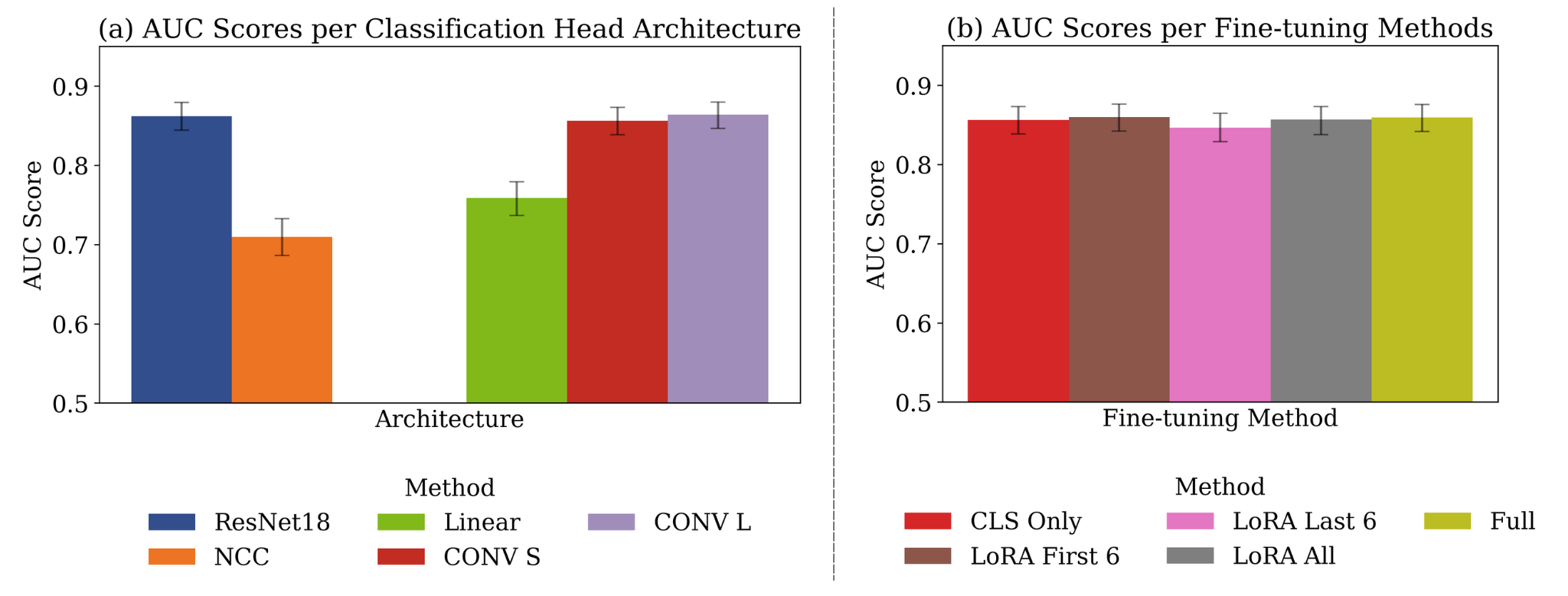}
\caption{(a) AUC per classification head architecture for each dataset. ResNet18 and NCC are included as a reference, (b) AUC per fine-tuning method. Error bars show the 95\% CI obtained by bootstrapping on the test set. } \label{fig:exp1_exp2}
\end{figure}

In this experiment, the SAM-Med3D encoder is frozen and FedAvg is used for aggregation. Classification performance is reported in Fig. \ref{fig:exp1_exp2} (a) and efficiency metrics are reported in Table \ref{tab:efficiency}. 

Training the classifier significantly outperforms the NCC (AUC = 0.71, 95\% CI: 0.69–0.73). Both convolutional heads significantly outperform the linear classifier (AUC = 0.76, 95\% CI: 0.74–0.78), with ``CONV S'' and ``CONV L'' achieving similar performance (AUC = 0.86, 95\% CI: 0.84–0.87 vs. 0.86, 95\% CI: 0.85–0.88), both matching ResNet18 (AUC = 0.86, 95\% CI: 0.84–0.88) while using <13\% of its parameters and 75\% of its FLOPs. Larger heads increase message size (0.5 kB for linear vs. 36 kB for ``CONV L'') and latency (1.3 ms for linear vs. 3.0 ms for ``CONV L''); and minimally impact computational efficiency. ``CONV S'' offers the best trade-off, retaining high AUC with lower communication cost.

\subsection{Fine-tuning Method}
This experiment uses ``CONV S'' as a classifier and FedAvg for aggregation. Diagnostic performance is shown in Fig. \ref{fig:exp1_exp2} (b) and efficiency metrics in Table \ref{tab:efficiency}. 

None of the fine-tuning methods yielded higher performance than that obtained by linear probing (AUC = 0.86, 95\% CI: 0.84–0.87). No significant differences are observed between LoRA configurations: LoRA All (AUC = 0.86, 95\% CI: 0.84–0.87), LoRA First 6 (AUC = 0.86, 95\% CI: 0.84–0.88), and LoRA Last 6 (AUC = 0.85, 95\% CI: 0.83–0.86). Full-tuning achieves a similar performance (AUC = 0.86, 95\% CI: 0.84-0.88) despite the larger number of trained parameters (92M). Fine-tuning the encoder increases computational cost compared to linear probing due to higher parameter counts and gradient computations. 

% Tuning shallower layers (LoRA first 6) requires more resources than tuning the last layers (GPU memory: 40 GB Avs. 26 GB), as gradients must propagate through the entire backbone.

\subsection{Federated Aggregation Technique}

\begin{figure}[t]
\includegraphics[width=\textwidth]{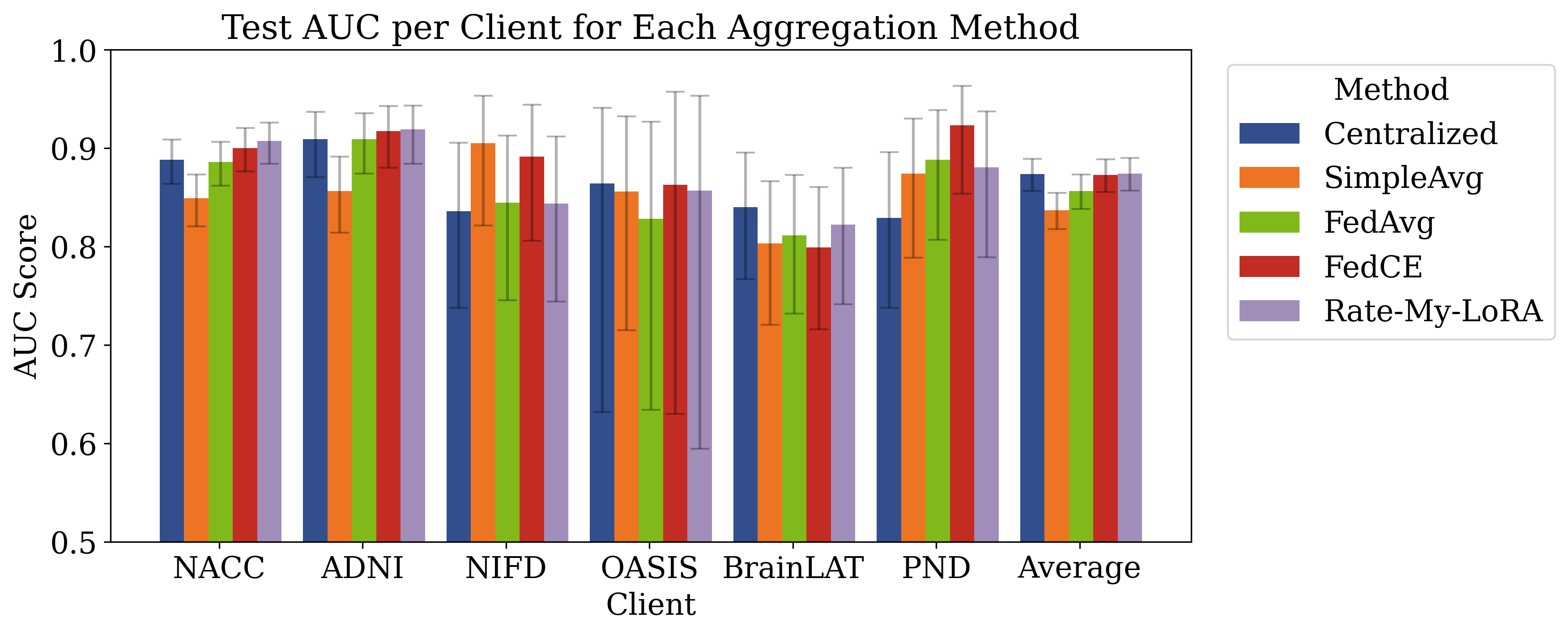}
\vspace{-0.7cm}
\caption{Test AUC score per client with different federated aggregation methods. Error bars show the 95\% CI (test set bootstrapping).} \label{fig:exp3}
\end{figure}

Fig. \ref{fig:exp3} shows the performance per client for each aggregation method, alongside the results of centralized training for comparison. The aggregation weights obtained with each method per round are presented in Fig. S.2 in the Appendix. 

Across the entire test set, both Rate-My-LoRA and FedCE match the performance of centralized training, with an AUC of 0.87 (95\% CI: 0.86–0.89). These methods slightly outperform FedAvg (AUC = 0.86, 95\% CI: 0.84–0.87) and significantly outperform simple averaging (AUC = 0.84, 95\% CI: 0.82–0.85). At the client level, ADNI and NACC show the highest gain with advanced aggregation methods, particularly when compared to simple averaging. Although not statistically significant, FedCE outperforms Rate-My-LoRA on PND (AUC = 0.92 vs. 0.88) and NIFD (AUC = 0.89 vs. 0.84). In particular, BrainLAT, which exhibits a slightly different intensity distribution, consistently yields lower AUCs across methods and does not benefit from advanced aggregation strategies.
\begin{table}[t]
\centering
\caption{Efficiency metrics for experiments on classification head (CLS) architecture and fine-tuning technique. Efficiency for CLS is assessed with the CLS-only finetuning setting, efficiency for fine-tuning techniques with the ``CONV S'' classification head.} % For number of trainable parameters, CLS refers to the CONV S parameters.}
\setlength{\tabcolsep}{3pt} 
\label{tab:efficiency}
\begin{tabular}{l|cccccc}
\toprule
Experiment & \makecell{Trainable\\Params (p)} & \makecell{Message\\Size (kB)} & \makecell{Latency\\(ms)} & \makecell{GPU Mem\\(GB)} & \makecell{Energy\\(MJ)} & \makecell{FLOPs\\(G)} \\
\midrule
ResNet18 & 33M & 232 & 6.1 & 44 & 2.5 & 251 \\
NCC & 0 & 0.5 & 1.2 & 14 & 0.2 & 184 \\
\midrule
Linear & 770 & 0.5 & 1.3 & 14 & 0.4 & 184 \\
CONV S & 1.7M & 16 & 1.9 & 14 & 2.9 & 185 \\
CONV L & 4.2M & 36 & 3.0 & 14 & 2.9 & 186 \\
\midrule
All & 92M + 1.7M & 424 & 6.9 & 44 & 4.1 & 185 \\
LoRA ALL & 294k + 1.7M & 19 & 2.0 & 40 & 4.0 & 186 \\
LoRA First 6 & 147k + 1.7M & 18 & 2.0 & 40 & 4.0 & 185 \\
LoRA Last 6 & 147k + 1.7M  & 18 & 2.0 & 26 & 3.8 & 185 \\
\bottomrule
\end{tabular}
\end{table}

\section{Discussion}
In this work, we implemented a framework for a systematic evaluation of federated FM fine-tuning for dementia classification using T1-weighted MRI. We investigated three key design choices and their impact on model performance and efficiency, leveraging a large dataset consisting of 6 different cohorts.

Our results show that FL enables effective fine-tuning of the SAM-Med3D segmentation model for a classification task, achieving comparable performance to conventional CNNs with higher efficiency. In addition, it approaches the performance of centralized fine-tuning, underscoring the promise of integrating FL with FMs for AI-based dementia diagnosis.
Our findings indicate that the classification head architecture has a substantial impact on performance. Incorporating convolutional layers on top of SAM-Med3D enhances performance by adapting its segmentation features to the classification task. We find that fine-tuning the encoder provides no significant performance improvement over using it as a frozen backbone, highlighting the high quality of the pre-trained features. This is especially valuable in FL, as freezing the backbone greatly reduces communication overhead without compromising performance. 
We observe that advanced aggregation strategies improve overall diagnostic performance compared to conventional methods. However, a more detailed analysis reveals that the gain is primarily driven by clients with large datasets (e.g. NACC), while smaller clients see limited benefit. This suggests that relying solely on validation-based client weighting may be insufficient. Supporting this, we find that FedCE, which incorporates gradient information, outperforms Rate-My-LoRA, which depends exclusively on validation metrics in smaller clients.
While real-world federations often involve heterogeneous hardware, we employ a homogeneous setup to minimize variability stemming from infrastructure differences. This controlled approach enables a more precise evaluation of how design choices influence efficiency.

While our study introduces a flexible framework for MRI-based dementia diagnosis with a broader applicability to other medical imaging tasks, it has a number of limitations. First, the evaluation is constrained by the scarcity of open-source 3D FMs for MRI data, limiting our experiments to SAM-Med3D and a selected set of design choices. As the field progresses and more models become available, future work will benchmark alternative FMs to determine their suitability for FL environments. Additionally, while we focus on linear probing and LoRA as the current de facto approaches in PEFT, emerging FL-specific methods could further optimize performance and communication efficiency. Exploring these techniques will be critical as FMs gain traction in FL. Finally, a deeper theoretical analysis of aggregation methods is essential. While the evaluated methods rely mostly on validation performance, integrating fairness-aware aggregation, convergence guarantees, and client-specific bias mitigation could improve both the performance and equity of the resulting models.

In essence, this work investigates federated fine-tuning of FMs within real-world, multi-source datasets, moving beyond simulated federated data to ensure clinical relevance and practical viability. By investigating foundational yet underexplored components of the federated fine-tuning paradigm, it lays the ground for broader and more in-depth future evaluations.

\begin{credits}
\subsubsection{\ackname}
This project is supported by a 2022 Erasmus MC Fellowship. Esther E. Bron is recipient of TAP-dementia, a ZonMw funded project (\#10510032120003). Esther E. Bron and Stefan Klein are recipients of EUCAIM, Cancer Image Europe, co-funded by the European Union under Grant Agreement 101100633. Data used in this study was partially obtained from the National Alzheimer’s Coordinating Center (NACC) database. MRI imaging data are part of the SCAN initiative. The NACC database is funded by NIA/NIH Grant U24 AG072122. SCAN was funded as a U24 grant (AG067418).

\subsubsection{\discintname}
The authors have no competing interests to declare that are
relevant to the content of this article.
\end{credits}
%
% ---- Bibliography ----
%
\bibliographystyle{splncs04}
\bibliography{references}

\appendix
\section{Appendix}

\begin{figure}[ht]
\centering
\includegraphics[width=\linewidth]{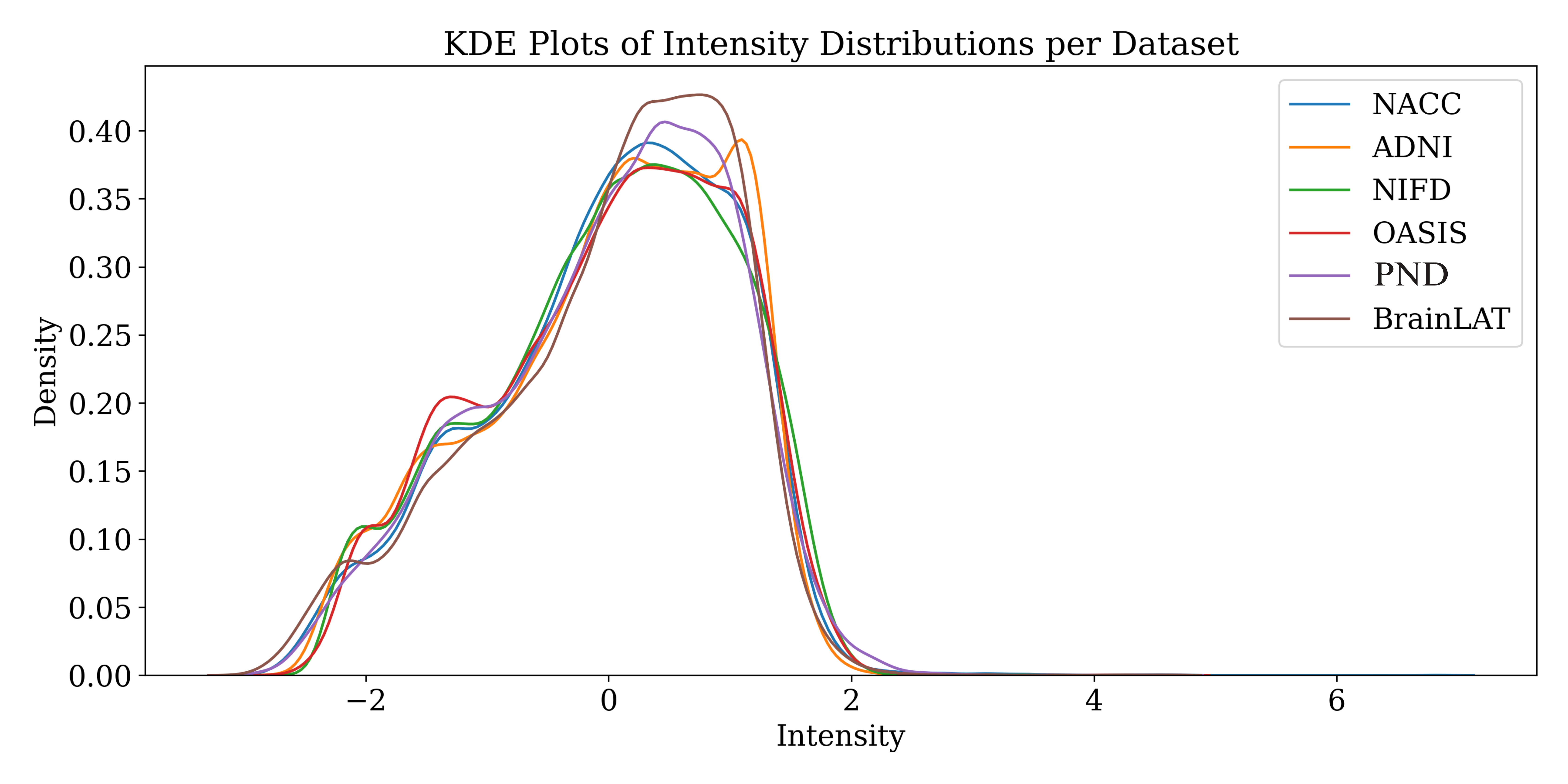} % Replace with your actual image
\caption{Intensity profiles per client after normalization, obtained using Kernel Density Estimation over all voxels within the brain mask from all scans.}
\label{fig:app1}
\end{figure}
\begin{figure}[h]
\centering
\includegraphics[width=\linewidth]{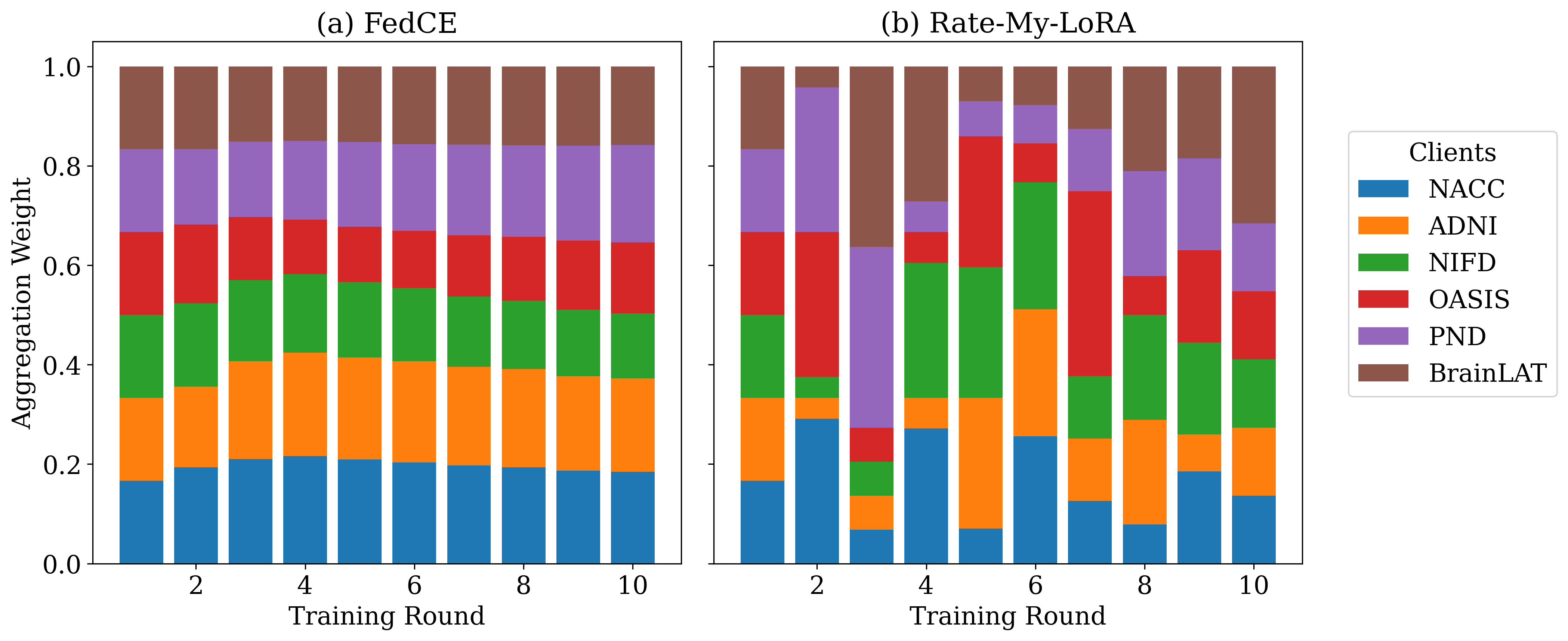} % Replace with your actual image
\caption{Aggregation weights assigned to each client per round by (a) FedCE and (b) Rate-My-LoRA. FedCE produces stable weight trajectories due to its use of running aggregation which incorporates past weights into current updates, with NACC and ADNI consistently contributing the most. The weights of smaller clients (e.g. PND) gradually increase, suggesting that FedCE progressively integrates their contributions as training evolves. In contrast, Rate-My-LoRA shows highly variable weighting across rounds, with several clients intermittently receiving negligible weights. This reflects Rate-My-LoRA’s reliance on per-round validation performance, which can lead to unstable client weighting and potential underutilization of data from smaller sites.}
\label{fig:app2}
\end{figure}

\begin{landscape}
\begin{table}[ht]
\centering
\caption{AUC scores per client across all experiments, reported as average [95\% CI] with bootstrapping on the test set. The highest average AUC for each client and experiment is highlighted in bold. RML: Rate-My-LoRA.}
\begin{tabularx}{\linewidth}{l*{7}{>{\centering\arraybackslash}X}}
\toprule
Method & ADNI & BrainLAT & NACC & NIFD & OASIS & PND & All \\
\midrule\multicolumn{8}{c}{\textit{Baselines}} \\
\midrule
ResNet18 & 0.93 [0.90, 0.96] & 0.82 [0.74, 0.88] & 0.88 [0.86, 0.91] & 0.79 [0.69, 0.88] & 0.88 [0.76, 0.94] & 0.83 [0.73, 0.90] & 0.86 [0.84, 0.88] \\
NCC &  0.74 [0.69, 0.79] & 0.72 [0.63, 0.80] & 0.72 [0.68, 0.75] & 0.74 [0.62, 0.83] & 0.79 [0.64, 0.89] & 0.74 [0.63, 0.83] & 0.71 [0.69, 0.73] \\
Centralized & 0.91 [0.87, 0.94] & 0.84 [0.77, 0.90] & 0.89 [0.86, 0.91] & 0.84 [0.74, 0.91] & 0.86 [0.63, 0.94] & 0.83 [0.74, 0.90] & 0.87 [0.86, 0.89] \\
\midrule
\multicolumn{8}{c}{\textit{Classification Head Architecture}} \\
\midrule
Linear & 0.81 [0.76, 0.85] & 0.74 [0.65, 0.82] & 0.79 [0.76, 0.82] & 0.84 [0.74, 0.91] & 0.82 [0.62, 0.93] & 0.73 [0.62, 0.82] & 0.76 [0.74, 0.78] \\
CONV S & \textbf{0.91} [0.87, 0.94] & 0.81 [0.73, 0.87] & 0.89 [0.86, 0.91] & 0.84 [0.75, 0.91] & 0.83 [0.63, 0.93] & \textbf{0.89} [0.81, 0.94] & \textbf{0.86} [0.84, 0.87] \\
CONV L & 0.90 [0.86, 0.93] & \textbf{0.83} [0.75, 0.89] & \textbf{0.90} [0.88, 0.92] & \textbf{0.87} [0.78, 0.93] & \textbf{0.86} [0.69, 0.95] & 0.86 [0.77, 0.92] & \textbf{0.86} [0.85, 0.88] \\
\midrule
\multicolumn{8}{c}{\textit{Fine-tuning Method}} \\
\midrule
All & \textbf{0.91} [0.87, 0.93] & \textbf{0.82} [0.73, 0.88] & 0.88 [0.86, 0.90] & \textbf{0.92} [0.84, 0.96] & 0.80 [0.59, 0.91] & 0.83 [0.73, 0.90] & \textbf{0.86} [0.84, 0.88] \\
CLS Only & \textbf{0.91} [0.87, 0.94] & 0.81 [0.73, 0.87] & 0.89 [0.86, 0.91] & 0.84 [0.75, 0.91] & 0.83 [0.63, 0.93] & \textbf{0.89} [0.81, 0.94] & \textbf{0.86} [0.84, 0.87] \\
LoRA All & 0.90 [0.86, 0.93] & \textbf{0.82} [0.74, 0.87] & 0.89 [0.87, 0.91] & 0.84 [0.75, 0.91] & \textbf{0.87} [0.71, 0.94] & 0.87 [0.77, 0.93] & \textbf{0.86} [0.84, 0.87] \\
LoRA first 6 & \textbf{0.91} [0.87, 0.93] & 0.81 [0.72, 0.87] & \textbf{0.90} [0.88, 0.92] & 0.83 [0.73, 0.90] & 0.85 [0.67, 0.94] & 0.83 [0.73, 0.90] & \textbf{0.86} [0.84, 0.88] \\
LoRA last 6 & 0.89 [0.85, 0.92] & 0.79 [0.70, 0.85] & 0.89 [0.87, 0.91] & 0.86 [0.77, 0.92] & 0.80 [0.57, 0.91] & 0.83 [0.73, 0.91] & 0.85 [0.83, 0.86] \\
\midrule
\multicolumn{8}{c}{\textit{Aggregation Technique}} \\
\midrule
SimpleAvg & 0.86 [0.81, 0.89] & 0.80 [0.72, 0.87] & 0.85 [0.82, 0.87] & \textbf{0.90} [0.82, 0.95] & \textbf{0.86} [0.71, 0.93] & 0.87 [0.79, 0.93] & 0.84 [0.82, 0.85] \\
FedAvg & 0.91 [0.87, 0.94] & 0.81 [0.73, 0.87] & 0.89 [0.86, 0.91] & 0.84 [0.75, 0.91] & 0.83 [0.63, 0.93] & 0.89 [0.81, 0.94] & 0.86 [0.84, 0.87] \\
FedCE & \textbf{0.92} [0.88, 0.94] & 0.80 [0.72, 0.86] & 0.90 [0.88, 0.92] & 0.89 [0.81, 0.94] & \textbf{0.86} [0.63, 0.96] & \textbf{0.92} [0.85, 0.96] & \textbf{0.87} [0.86, 0.89] \\
RML & \textbf{0.92} [0.88, 0.94] & \textbf{0.82} [0.74, 0.88] & \textbf{0.91} [0.88, 0.93] & 0.84 [0.74, 0.91] & \textbf{0.86} [0.59, 0.95] & 0.88 [0.79, 0.94] & \textbf{0.87} [0.86, 0.89] \\

\bottomrule
\end{tabularx}
\label{tab:auc_results}
\end{table}
\end{landscape}

\end{document}